\documentclass{article}
\usepackage{spconf,amsmath,graphicx}

\usepackage{xcolor}
\usepackage{enumerate}
\usepackage{diagbox}
\usepackage{enumitem}
\usepackage{multirow}
\usepackage{comment}
\usepackage{lipsum}
\usepackage{booktabs}
\usepackage{subcaption}
\usepackage[super]{nth}

\title{AIPNet: Generative Adversarial Pre-training of Accent-Invariant Networks for End-to-End Speech Recognition}
\name{Yi-Chen Chen$^{1,2}$, Zhaojun Yang$^1$, Ching-Feng Yeh$^1$, Mahaveer Jain$^1$, Michael L. Seltzer$^1$}
\address{$^1$Facebook AI, USA\\
$^2$Speech Processing and Machine Learning Laboratory, National Taiwan University, Taiwan
}

\begin{document}
\ninept
\maketitle
\begin{abstract}
As one of the major sources in speech variability, accents have posed a grand challenge to the robustness of speech recognition systems. In this paper, our goal is to build a unified end-to-end speech recognition system that generalizes well across accents. For this purpose, we propose a novel pre-training framework \texttt{AIPNet} based on generative adversarial nets (GAN) for accent-invariant representation learning: \textbf{A}ccent \textbf{I}nvariant \textbf{P}re-training \textbf{Net}works. We pre-train \texttt{AIPNet} to disentangle accent-invariant and accent-specific characteristics from acoustic features through adversarial training on accented data for which transcriptions are not necessarily available. We further fine-tune \texttt{AIPNet} by connecting the accent-invariant module with an attention-based encoder-decoder model for multi-accent speech recognition. In the experiments, our approach is compared against four baselines including both accent-dependent and accent-independent models.
Experimental results on $9$ English accents show that the proposed approach outperforms all the baselines by $2.3\sim4.5\%$ relative reduction on average WER when transcriptions are available in all accents and by $1.6\sim6.1\%$ relative reduction when transcriptions are only available in US accent. 
\end{abstract}
\begin{keywords}
Generative adversarial network, end-to-end speech recognition, accent-invariance
\end{keywords}
\section{Introduction}
\label{sec:intro}

Accents are defined as variations in pronunciation within a language and are often peculiar to geographical regions, individuals, social groups, etc. As one of the major sources in speech variability, accents have posed a grand technical challenge to the robustness of ASR systems. Due to the acoustic discrepancy among accents, an ASR system that is trained on the speech data of one accent (e.g., native) often fails to recognize speech of another unseen accent (e.g., non-native). In this work, we focus on learning accent-invariant representations, aiming to build a universal ASR system that is generalizable across accents.

There is an extensive literature on multi-accent modeling for speech recognition \cite{rao2017multi} \cite{lin2009study}. The existing approaches can be categorized into two classes in general: accent-independent and accent-dependent. Accent-independent modeling focuses on building a universal model that generalizes well across accents. One popular baseline is to train a model on all the data of different accents \cite{elfeky2016towards} \cite{kamper2011multi} \cite{vergyri2010automatic}. Elfeky \emph{et al.}~have attempted to build a unified multi-accent recognizer from a pre-defined unified set of CD states by learning from the ensemble of accent-specific models \cite{elfeky2016towards}. Yang \emph{et al.}~have proposed to jointly model ASR acoustic model and accent identification classifier through multi-task learning \cite{yang2018joint}. Accent-dependent approaches either take accent-related information, such as accent embedding or i-vectors, as an complementary input in the modeling or adapt a unified model on accent-specific data \cite{li2018multi} \cite{chen2015improving} \cite{yoo2019highly} \cite{vu2014multilingual} \cite{huang2014multi}. Accent-dependent models usually outperform the unified ones with known accent labels or on an accent-specific dataset, while accent-independent models demonstrate better generalizability on average when accent labels are unavailable during testing.


Generative adversarial nets (GAN) \cite{goodfellow2014generative} or gradient reverse technique \cite{ganin2014unsupervised} has gained popularity in learning a representation that is invariant to domains or conditions \cite{chen2018phonetic} \cite{serdyuk2016invariant} \cite{sun2018domain} \cite{bousmalis2017unsupervised}. Serdyuk \emph{et al.}~have applied adversarial training to generate noise-invariant representations for speech recognition \cite{serdyuk2016invariant}. Gradient reversal training has recently been used for learning domain-invariant features to alleviate the mismatch between accents during training \cite{sun2018domain}. Bousmalis \emph{et al.}~have proposed to a GAN-based pixel-level transformation from one domain to the other and have shown great improvement over state-of-the-art on a number of domain adaptation tasks \cite{bousmalis2017unsupervised}.

This work focuses on learning accent-invariance with the goal of building a unified accent-independent system for end-to-end speech recognition. Pre-training has shown its superiority in many computer vision and NLP tasks \cite{krizhevsky2012imagenet} \cite{devlin2018bert} \cite{mikolov2013distributed}, while research efforts on accent model pre-training thus far have been limited. We propose a novel pre-training framework \texttt{AIPNet} based on GAN for accent-invariant representation learning: \textbf{A}ccent \textbf{I}nvariant \textbf{P}re-training \textbf{Net}works. Unlike most of the existing work that unites the modeling of acoustics and accents in a single stage, our approach decouples accent modeling from acoustic modeling and consists of two stages: pre-training and fine-tuning. In the pre-training stage, \texttt{AIPNet} is built through adversarial training to disentangle accent-invariant and accent-specific characteristics from acoustic features. As transcriptions are not needed in pre-training, \texttt{AIPNet} allows us to make use of many possible accent resources for which transcriptions are unavailable. In the fine-tuning stage, we adopt an attention-based encoder-decoder model for sequence-to-sequence speech recognition. Specifically, we plug in the accent-invariant embeddings in \texttt{AIPNet} into ASR model for further optimization. Experimental results on $9$ English accents show significant WER reduction compared to four popular baselines, indicating the effectiveness of \texttt{AIPNet} on accent-invariance modeling. As a general framework for learning domain-invariance, \texttt{AIPNet} can be easily generalized to model any variabilities, such as speakers or speech noise, in addition to accents.

\section{AIPNet}
\label{sec:approach}

\begin{figure}[t]
  \centering
  \includegraphics[width=0.95\linewidth]{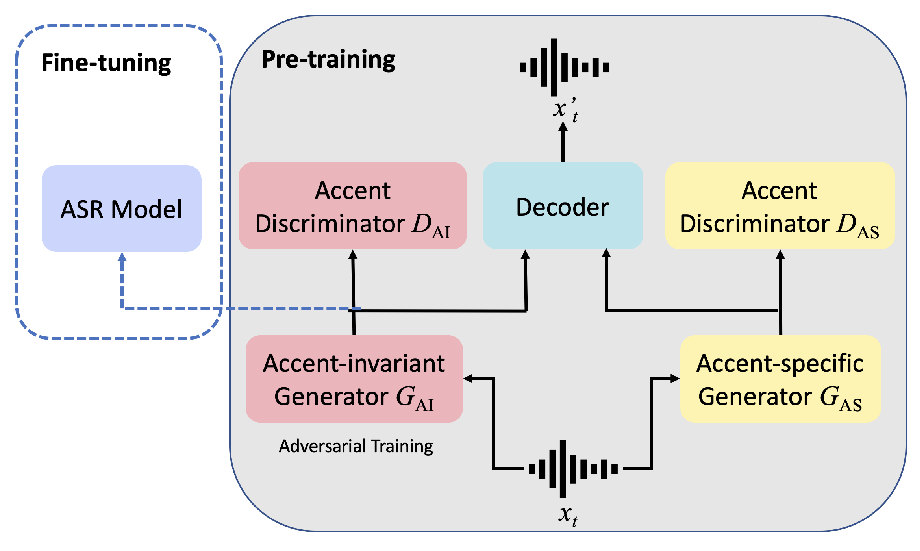}
 \caption{The framework of \texttt{AIPNet} including both pre-training and fine-tuning stages.}
  \label{fig:approach} 
\end{figure}

In this section we describe \texttt{AIPNet} in details. Our approach  consists of two stages: pre-training and fine-tuning. In the pre-training stage, the model is built through adversarial training with the goal of learning accent-invariant representations. In the fine-tuning stage, we stack the pre-trained model with downstream tasks for further optimization. In this work, we use end-to-end ASR as a downstream application, focusing on improving accent robustness for speech recognition. The framework of \texttt{AIPNet} is illustrated in Fig.~\ref{fig:approach}.


Suppose the input is an utterance $\mathbf{X} = (\mathbf{x}_1, \mathbf{x}_2, ..., \mathbf{x}_T)$, where $\mathbf{x}_t$ represents the feature vector at time step $t$. The speaker accent corresponding to $\mathbf{x}_t$ is denoted as $a_t \in \{1, 2, ..., C\}$, where $C$ is the number of accents in the training data.


\subsection{Accent-Invariance Pre-training}
The goal of pre-training is to learn accent-invariant representations from accented training data. We define three types of losses for this purpose, including adversarial loss to disentangle accent-invariant and accent-specific information, reconstruction loss to enforce acoustic characteristics to be preserved in the disentangled representations, as well as consistency regularization to detach linguistic information from accent-specific representations.
 
\subsubsection{Adversarial Loss}
\label{subsec:adversarial}
To learn accent-invariant representations, we define two mappings from speech data: accent-invariant generator $G_{AI}(\mathbf{x}_t)$ and accent-specific generator $G_{AS}(\mathbf{x}_t)$. We also define two discriminators $D_{AI}(G_{AI})$ and $D_{AS}(G_{AS})$ that output probabilities of accents to ensure that $G_{AI}$ and $G_{AS}$ encode the corresponding information. Specifically, we train $D_{AI}$ and $D_{AS}$ to maximize the probability of assigning correct accent labels to samples from $G_{AI}$ and $G_{AS}$ respectively, \emph{i.e.,} minimizing cross-entropy loss $L_{CE}^{AI}$ and $L_{CE}^{AS}$:
\begin{align}
\min_{D_{AS}, G_{AS}} L_{CE}^{AS} &= \sum_{t=1}^T -\log P(a_t | G_{AS}(\mathbf{x}_t)),\\
\min_{D_{AI}} L_{CE}^{AI} &= \sum_{t=1}^T -\log P(a_t | G_{AI}(\mathbf{x}_t)).
\label{eq:L_CE_AS}
\end{align}
To decouple accent-related information from $G_{AI}$, we simultaneously train $G_{AI}$ such that $D_{AI}$ is confused about accent labels of samples from $G_{AI}$. The objective is to maximize cross-entropy loss $L_{CE}^{AI}$, equivalent to minimize the negative cross-entropy:
\begin{align}
\min_{G_{AI}} -L_{CE}^{AI} = \sum_{t=1}^T \log P(a_t | G_{AI}(\mathbf{x}_t)).
\label{eq:L_NE_AI}
\end{align}

\subsubsection{Reconstruction Loss}
\label{subsec:autoencoding}

The adversarial loss defined between $D_{AI}$ and $G_{AI}$ enforces that accent-specific information is disentangled from $G_{AI}$ but preserved in $G_{AS}$. To ensure acoustics characteristics are encoded in the representations from both generators, we further define a decoder with autoencoding structure to reconstruct speech feature $\mathbf{x}_t$ from $G_{AI}(\mathbf{x}_t)$ and $G_{AS}(\mathbf{x}_t)$. The decoder is trained by minimizing the reconstruction error $L_{R}$:
\begin{align}
\min_{decoder, G_{AI}, G_{AS}} L_{R} = \sum_{t=1}^T \parallel {\mathbf{x}'_t} - {\mathbf{x}_t} \parallel_2^2.
\label{eq:L_R}
\end{align}

\subsubsection{Consistency Regularization}
\label{subsec:consistency}
Accent-specific attributes are generally stable within an utterance while linguistic-related acoustics have larger intra-utterance variance across time frames. Inspired by the utterance-level stability of accent-specific attributes, we impose a consistency regularization for $G_{AS}(\mathbf{x}_t)$ such that accent-specific representations from $G_{AS}$ are consistent across time frames within an utterance:
\begin{align}
\min_{G_{AS}} L_{CR} = \sum_{t=1}^{T-1} \parallel {G_{AS}(\mathbf{x}_{t+1}) - G_{AS}(\mathbf{x}_{t})} \parallel_2^2.
\label{eq:L_CR}
\end{align}
This regularization reinforces the preservation of accent-specific information in $G_{AS}$ meanwhile implicitly encourages linguistic content to be disentangled from $G_{AS}$. The multi-scale nature of information in speech data has also been applied in voice conversion and speech denoising \cite{hsu2017unsupervised}.

\subsubsection{Iterative Training}
\label{subsec:iterative}
Given the minmax two-player game between $D_{AI}$ and $G_{AI}$, \texttt{AIPNet} pre-training is designed of repeating the following two steps in an iterative manner.
\begin{itemize}
\item Update the discriminator $D_{AI}$ by minimizing $L_D$,
\item Freeze the discriminator $D_{AI}$ and update the rest of the network by minimizing $L_G$,
\end{itemize}
\vspace{-6pt}
\begin{align}
L_D &= L_{CE}^{AI}, \\
L_G &= -L_{CE}^{AI} + \lambda_1 L_{CE}^{AS} + \lambda_2 L_{R} + \lambda_3 L_{CR},
\end{align}
where $\lambda$s are hyper-parameters.


\begin{table*}[t]
\centering
\caption{WER (\%) of different approaches in each accent in supervised setting. \texttt{F1} indicates fine-tuning with $L_{ASR}$; \texttt{F2} indicates fine-tuning with $L'_G$; AI indicates accent-independent model; AD indicates accent-dependent model.}
\label{table:supervised}
\begin{tabular}{c|c c|c|c c|c c c c c c c c}
\toprule[2pt]

\multicolumn{3}{c|}{Approach} &
\multicolumn{1}{c|}{Ave.} &
\multicolumn{1}{c}{US} &
\multicolumn{1}{c|}{Non-US} &
\multicolumn{1}{c}{CA} &
\multicolumn{1}{c}{FR} &
\multicolumn{1}{c}{IN} &
\multicolumn{1}{c}{KR} &
\multicolumn{1}{c}{PH} &
\multicolumn{1}{c}{LA} &
\multicolumn{1}{c}{GB} &
\multicolumn{1}{c}{VN} \\
\toprule[2pt]

\multirow{4}{*}{Baselines} &
\texttt{B1} & AI & 8.7 & 5.7 & 9.0 & 6.4 & 8.4 & 11.2 & \bf{9.9} & 7.2 & \bf{7.8} & 8.0 & 13.0 \\ \cline{2-14}
 & \texttt{B2} & AI  & 8.8 & \bf{5.0} & 9.1 & 6.6 & 9.3 & 11.0 & 10.3 & \bf{6.7} & 8.1 & 8.1 & 12.9 \\ \cline{2-14}
 & \texttt{B3} & AD & 8.6 & 5.4 & 8.9 & 6.7 & 8.5 & 10.9 & 10.0 & 6.8 & 8.6 & 7.9 & \bf{12.0} \\ \cline{2-14}
 & \texttt{B4} & AI & 8.8 & 5.8 & 9.1 & 6.1 & 8.5 & 11.7 & 10.7 & 7.4 & 8.4 & \bf{7.8} & \bf{12.0} \\ 
 \toprule[1.5pt]
\multirow{2}{*}{\texttt{AIPNet}} & 
\texttt{F1} & AI & \bf{8.4} & 5.6 & \bf{8.7} & \bf{6.0} & \bf{8.1} & \bf{9.9} & 10.3 & 6.9 & 8.0 & \bf{7.8} & 12.4 \\ \cline{2-14}
& \texttt{F2} & AI & 10.1 & 6.2 & 10.5 & 7.9 & 10.1 & 12.8 & 12.1 & 8.2 & 9.5 & 9.4 & 13.9 \\ 
\toprule[2pt]
\end{tabular}
\end{table*}

\begin{table*}[t]
\centering
\caption{WER (\%) of different approaches in each accent in semi-supervised setting. \texttt{F1} indicates fine-tuning with $L_{ASR}$; \texttt{F2} indicates fine-tuning with $L'_G$; PL indicates pseudo labeling; AI indicates accent-independent model; AD indicates accent-dependent model.}
\label{table:semi-supervised}
\begin{tabular}{c|c|c c|c|c c|c c c c c c c c}
\toprule[2pt]

\multicolumn{4}{c|}{Approach} &
\multicolumn{1}{c|}{Ave.} &
\multicolumn{1}{c}{US} &
\multicolumn{1}{c|}{Non-US} &
\multicolumn{1}{c}{CA} &
\multicolumn{1}{c}{FR} &
\multicolumn{1}{c}{IN} &
\multicolumn{1}{c}{KR} &
\multicolumn{1}{c}{PH} &
\multicolumn{1}{c}{LA} &
\multicolumn{1}{c}{GB} &
\multicolumn{1}{c}{VN} \\
\toprule[2pt]

 \multirow{2}{*}{w/o PL} & Baseline &  
 \texttt{B1} & AI & 29.9 & 10.6 & 31.8 & 22.0 & 33.1 & 41.0 & 33.1 & 28.2 & 28.7 & 28.3 & 40.6 \\ \cline{2-15}
 & \texttt{AIPNet} & \texttt{F1} & AI & 27.9 & \bf{9.4} & 29.8 & 20.1 & 30.8 & 39.0 & 32.8 & 25.5 & 26.4 & 25.3 & 39.2 \\
 \toprule[1.5pt]
\multirow{4}{*}{w/ PL} & \multirow{4}{*}{Baselines} & 
\texttt{B1} & AI & 26.2 & 10.3 & 27.8 & 18.6 & 28.3 & 36.1 & 29.6 & 24.8 & 25.1 & 24.4 & 35.8 \\ \cline{3-15}
 & & \texttt{B2} & AI & 25.9 & \bf{9.4} & 27.6 & 19.0 & 27.7 & 36.5 & 29.6 & 24.2 & 23.8 & 25.1 & 34.9 \\ \cline{3-15}
 & & \texttt{B3} & AD & 25.9 & 9.6 & 27.5 & 19.5 & 28.0 & 36.4 & 29.1 & 23.7 & 24.2 & 24.8 & 35.0 \\ \cline{3-15}
 & & \texttt{B4} & AI & 25.0 & 9.7 & 26.5 & \bf{18.1} & 26.7 & 34.9 & 28.3 & 23.7 & 23.4 & 23.7 & 33.6 \\ 
 \toprule[1.5pt]
 \multirow{2}{*}{w/ PL}  & \multirow{2}{*}{\texttt{AIPNet}} & 
 \texttt{F1} & AI & 25.7 & 12.1 & 27.0 & 19.7 & 27.4 & 34.7 & 28.9 & 23.0 & 23.6 & 24.5 & 34.6 \\ \cline{3-15}
 & & \texttt{F2} & AI & \bf{24.6} & 11.8 & \bf{25.9} & 19.0 & \bf{26.0} & \bf{32.6} & \bf{28.0} & \bf{22.2} & \bf{22.8} & \bf{23.1} & \bf{33.5} \\
 \toprule[2pt]
\end{tabular}
\vspace{-0.1cm}
\end{table*}

\subsection{Fine-tuning for End-to-End Speech Recognition}
\label{subsec:ASR}
In the fine-tuning stage, the outputs of $G_{AI}$ which encode accent-invariant linguistic content can be plugged in as inputs of any downstream speech tasks that aim to improve accent robustness, as shown in Fig.~\ref{fig:approach}. In this work, we focus on multi-accent speech recognition and adopt Listen, attend and spell (\texttt{LAS}), a popular attention-based encoder-decoder model \cite{chan2016listen} for sequence-to-sequence speech recognition. \texttt{LAS} consists of an encoder encoding an input sequence into high-level representations as well as an attention-based decoder generating a sequence of labels from the encoded representations. The encoder is typically a unidirectional or bidirectional LSTM and the decoder is a unidirectional LSTM.

The label inventory for \texttt{LAS} modeling consists of $200$ word pieces and is further augmented with two special symbols $<$sos$>$ and $<$eos$>$ indicating the start of a sentence and the end of a sentence respectively. \texttt{LAS} models the posterior probability of a label sequence $\mathbf{y}$ given the input feature sequence $G_{AI}(\mathbf{X})$ and the previous label history $\mathbf{y}_{1:j-1}$:
\begin{align}
P(\mathbf{y} | G_{AI}(\mathbf{X})) = \prod_{j=1} P(y_j | G_{AI}(\mathbf{X}), \mathbf{y}_{1:j-1}).
\end{align}
Both encoder and decoder can be trained jointly for speech recognition by maximizing the log probability or minimizing $L_{ASR}$:
\begin{align}
L_{ASR} = \sum_{j=1} -\log P(y_j | G_{AI}(\mathbf{X}), \mathbf{y}_{1:j-1}).
\end{align}
There are two ways of fine-tuning: 1) fine-tune $G_{AI}$ and \texttt{LAS} with $L_{ASR}$. This requires only transcriptions in the training data; 2) continue with adversarial training as described in Sec \ref{subsec:iterative} with $L'_G = L_G + \lambda_4 L_{ASR}$. This requires both transcriptions and accent labels in the training data. In the experiments, we report results of using both ways.

\section{Experiments}
\label{sec:exp}

\subsection{Dataset}
\label{subsec:dataset}
The dataset used in experiments contains utterances in a variety of domains, such as weather or music, collected through crowdsourced workers. There are $9$ English accents in total in the dataset, including United States (US), Korea (KR), Philippines (PH), Canada (CA), India (IN), France (FR), Britain (GB), Vietnam (VN) and Latin America (LA). The training set contains $4$M ($3.8$K hours) utterances among which $1\%$ is split as validation data. Particularly, there are $1$M and $780$K utterances in US and LA respectively and about $330$K data in each of the remaining accents. The testing set has $10.8$K utterances with $1.2$K utterances in each accent. In both training and testing sets, we extract acoustic features using $80$-dimensional log Mel-filterbank energies that are computed over a $25$ms window every $10$ms.

\subsection{Experimental Setup}
\label{subsec:setup}
The architecture of each module in \texttt{AIPNet} is a multi-layer LSTM. Specifically, we represent $G_{AI}$, $G_{AS}$ and decoder using $2$ LSTM layers with a hidden size of $768$, $256$ and $1024$ respectively. $D_{AI}$ and $D_{AS}$ are represented by a LSTM layer with softmax outputs.
The configuration of \texttt{LAS} includes a $4$-layer LSTM encoder and a $2$-layer LSTM decoder, each with a hidden size of $1024$. The hyperparameters $(\lambda_1, \lambda_2, \lambda_3, \lambda_4)$ are swept within the range $[0.1, 30]$. Our experiments have shown that the final results are generally stable across different hyperparameter settings. For simplicity, we report results with $(\lambda_1, \lambda_2, \lambda_3, \lambda_4) = (1, 10, 10, 10)$ in this paper. We use batch size of $16,000$ tokens with $32$ GPUs for training. We use Adam with learning rate of $5 \times 10^{-4}$ in pre-training and $2.5 \times 10^{-4}$ in fine-tuning, $\beta_1 = 0.9$, $\beta_2 = 0.999$. A dropout rate of $0.1$ is applied to all the layers. We pre-train \texttt{AIPNet} for $15$ epochs and fine-tune \texttt{LAS} for $20$ epochs. During inference, speech features are fed into $G_{AI}$ that is absorbed as part of \texttt{LAS} encoder and outputs of \texttt{LAS} are decoded using beam size of $20$ without any external language model.


\subsection{Baselines}
We compare our approach against four popular baselines \texttt{B1}-\texttt{B4} for multi-accent speech recognition in the experiments. \texttt{B1} is an accent-independent model which is trained on the data from all the accents. \texttt{B2} and \texttt{B3} have shown strong performance on multi-accent speech recognition in \cite{li2018multi}. Specifically, we append accent labels at the end of each label sequence and \texttt{B2} is trained on the updated sequences from all accents. As accent information is not required at inference, \texttt{B2} is accent-independent. When training \texttt{B3} which is accent-dependent, we transform accent $1$-hot vector into an embedding through a learnable linear matrix and feed the learned embedding into \texttt{LAS} encoder. During \texttt{B1}-\texttt{B3} training, $G_{AI}$ is part of \texttt{LAS} encoder containing $6$ LSTM layers (see Section \ref{subsec:setup}). \texttt{B4} is the most similar to our approach in spirits, aiming to learn accent-invariant features through gradient reversal \cite{sun2018domain}. The gradient reversal approach keeps modules of $G_{AI}$, $D_{AI}$ and ASR model in Fig.~\ref{fig:approach}. Instead of using iterative training in Section \ref{subsec:iterative}, we add a gradient reversal layer between $G_{AI}$ and $D_{AI}$ to reverse the backpropagated gradient for $G_{AI}$ training. For more details about \texttt{B4}, we refer readers to \cite{sun2018domain}.

\begin{figure}[t]
\begin{subfigure}{.24\textwidth}
  \centering
  \includegraphics[width=.95\linewidth]{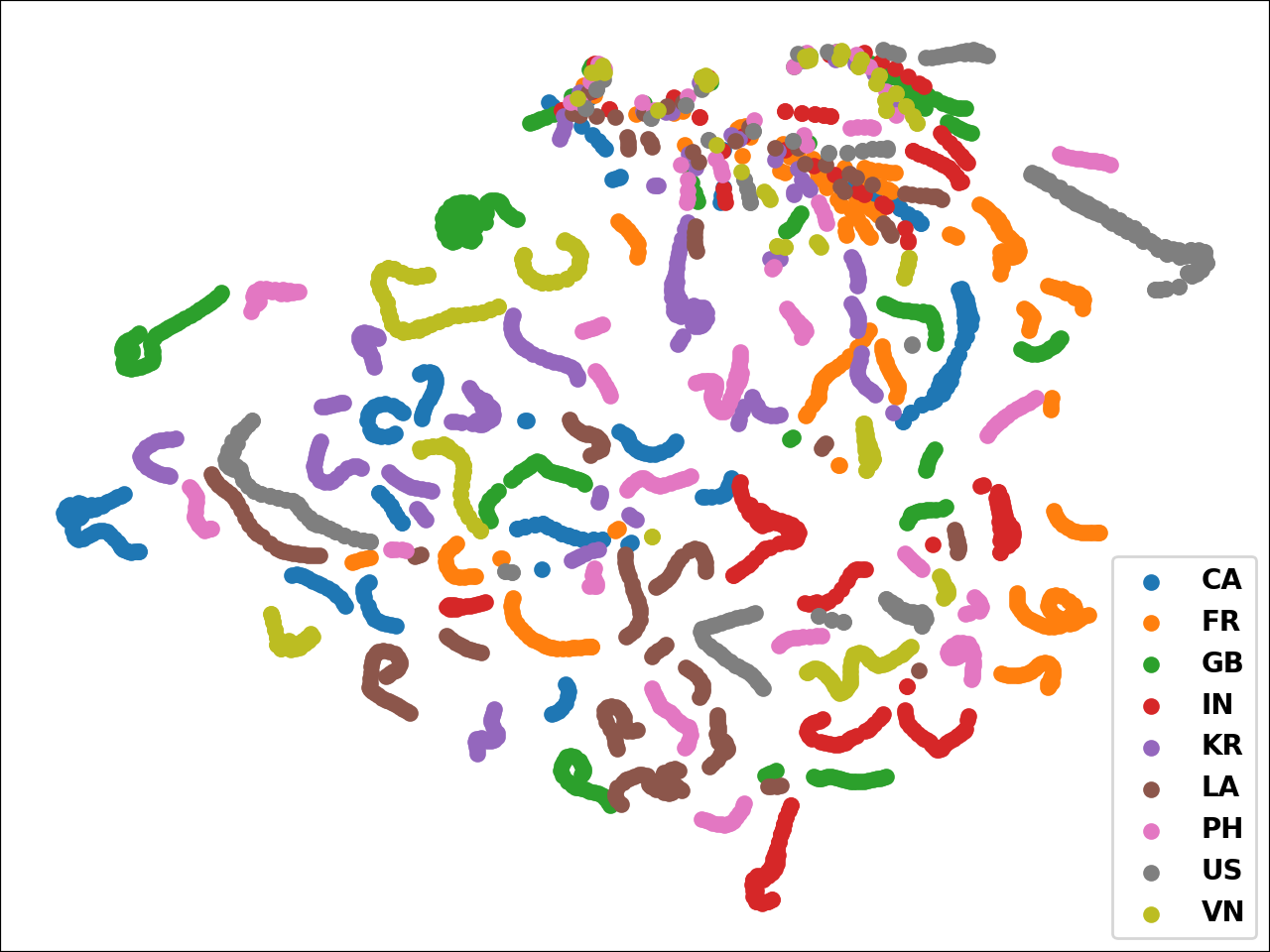}
  \caption{$G_{AI}$ embedding from B1.}
  \label{fig:tsne_B1}
\end{subfigure}%
\begin{subfigure}{.24\textwidth}
  \centering
  \includegraphics[width=.95\linewidth]{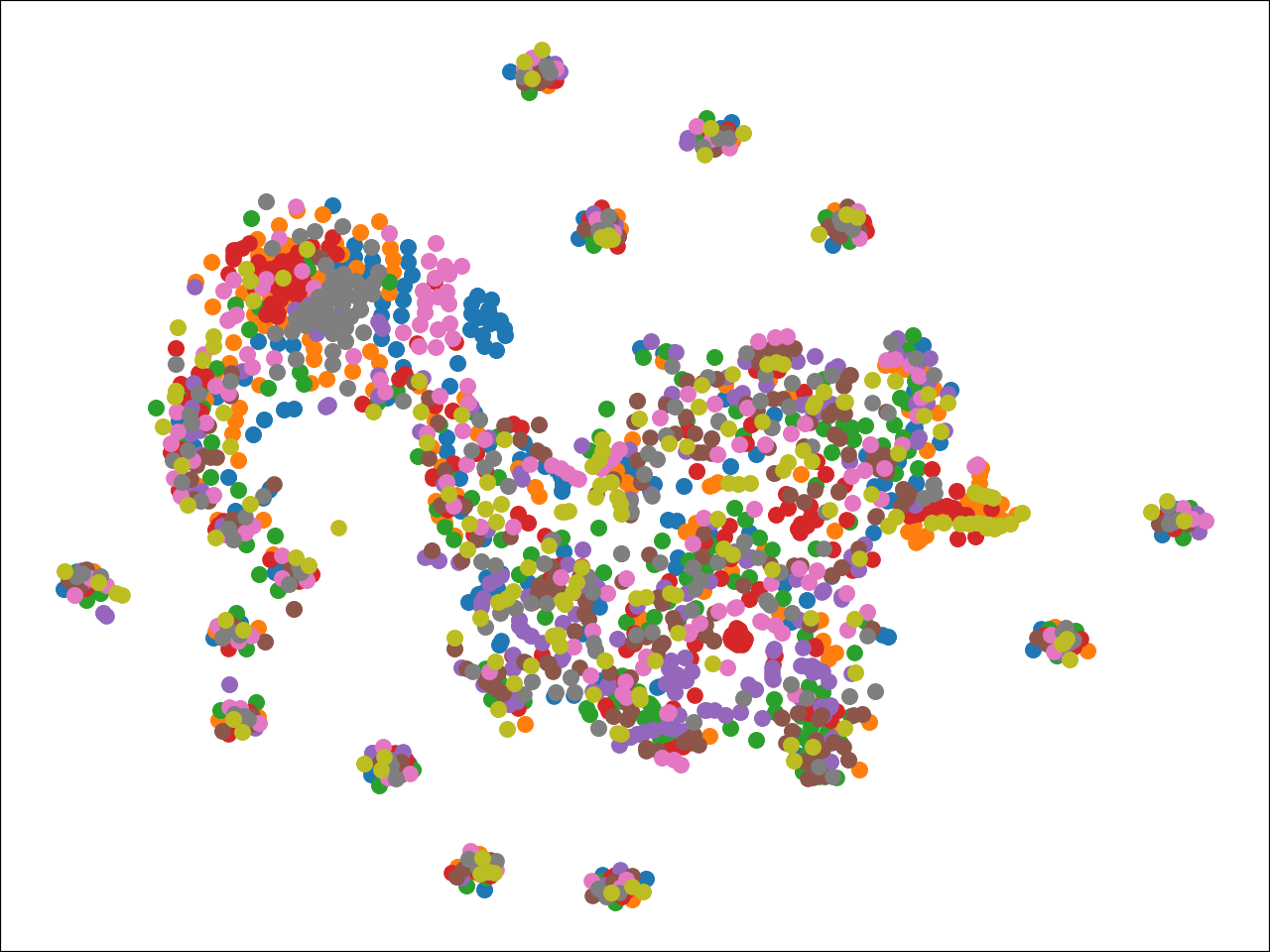}
  \caption{$G_{AI}$ embedding from F2.}
  \label{fig:tsne_F2}
\end{subfigure}
\caption{t-SNE 2-D plots of $G_{AI}$ embedding from B1 and F2 (w/ PL) in each accent. Each color represents each accent.}
\label{fig:tsne}
\end{figure}

\subsection{Experimental Results}
\label{subsec:results}
As described in Section \ref{sec:approach}, \texttt{AIPNet} pre-training requires only accent labels in the training data. This approach hence becomes especially useful when there is a large number of accented data without available transcriptions. We design experiments in two settings, \emph{i.e.,} supervised setting where transcriptions are available in all accents and semi-supervised setting where transcriptions are available only in US accent.

\subsubsection{Results in Supervised Setting}
\label{subsubsec:supervised}

Table~\ref{table:supervised} summarizes the results of different approaches in supervised setting. In our approach, we report results of fine-tuning $G_{AI}$ and \texttt{LAS} with $L_{ASR}$ using transcriptions (\texttt{F1}), as well as those of fine-tuning the entire network with $L'_G$ using both transcriptions and accent labels (\texttt{F2}). We can see that fine-tuning with $L_{ASR}$ (\texttt{F1}) outperforms the baselines by $2.3\sim4.5\%$ relative reduction on average WER. Compared to all the baselines, \texttt{F1} has achieved improvement in CA, FR, GB, and especially IN ($9.1\sim15.3\%$ reduction) but has shown a mediocre performance in each of the remaining accents. 



\subsubsection{Results in Semi-supervised Setting}
\label{subsubsec:semi-supervised}

In semi-supervised setting where transcriptions are available only in US accent, we compare the performance between \texttt{B1} and \texttt{F1}. The results are presented in the first two rows of Table \ref{table:semi-supervised}. As \texttt{B2}, \texttt{B3}, \texttt{B4} and \texttt{F2} require the availability of pairs of transcriptions and accent labels for training, the results of these approaches are not available in such scenario. The results have shown that our approach significantly outperforms the baseline model in all accents, achieving $3.4\sim11.3\%$ relative WER reduction. 

One popular and effective method for semi-supervised learning is to generate target pseudo labels for unlabeled data using an initial model \cite{lee2013pseudo}. To achieve better performance, we generate pseudo transcriptions for non-US training data using the US model. As a result, we are able to follow all the experiments in supervised setting in Section \ref{subsubsec:supervised}. The results with pseudo labeling (PL) are presented in the last six rows of Table \ref{table:semi-supervised}. By comparing the performance between models with and without pseudo labeling, we can observe that pseudo labeling has shown significant gains for all the approaches and almost in each accent, exhibiting its effectiveness on improving generalization performance using unlabeled data. In the scenario with pseudo labeling, fine-tuning with $L'_G$ (\texttt{F2}) outperforms the baselines by $1.6\sim6.1\%$ relative reduction on average WER and consistently achieves the best performance in all non-US accents except for CA.

\subsection{Analysis}
In this section, we analyze the properties of \texttt{AIPNet} to better understand its superiority for multi-accent speech recognition. Without loss of generality, we use \texttt{B1} and \texttt{F2} (w/ PL) in semi-supervised setting in the analysis.

\textbf{Learning accent-invariance} To comprehend the effectiveness of \texttt{AIPNet} on learning accent-invariant representations, we extract embedding (outputs) of $G_{AI}$ from \texttt{B1} and \texttt{F2} respectively for $300$ data samples in each accent. Fig.~\ref{fig:tsne} shows t-SNE $2$-D visualization of $G_{AI}$ embedding from \texttt{B1} (Fig.~\ref{fig:tsne_B1}) and \texttt{F2} (Fig.~\ref{fig:tsne_F2}) respectively for each accent \cite{maaten2008visualizing}. As can be seen, $G_{AI}$ outputs from the baseline \texttt{B1} tend to be clustered in each accent while those from our approach \texttt{F2} are mixed across different accents. The visualization demonstrates the validity of the accent-invariant features learned through \texttt{AIPNet} and further explains the better generalization performance that our approach has achieved across accents.

\begin{figure}[t]
  \centering
  \includegraphics[width=.85\linewidth]{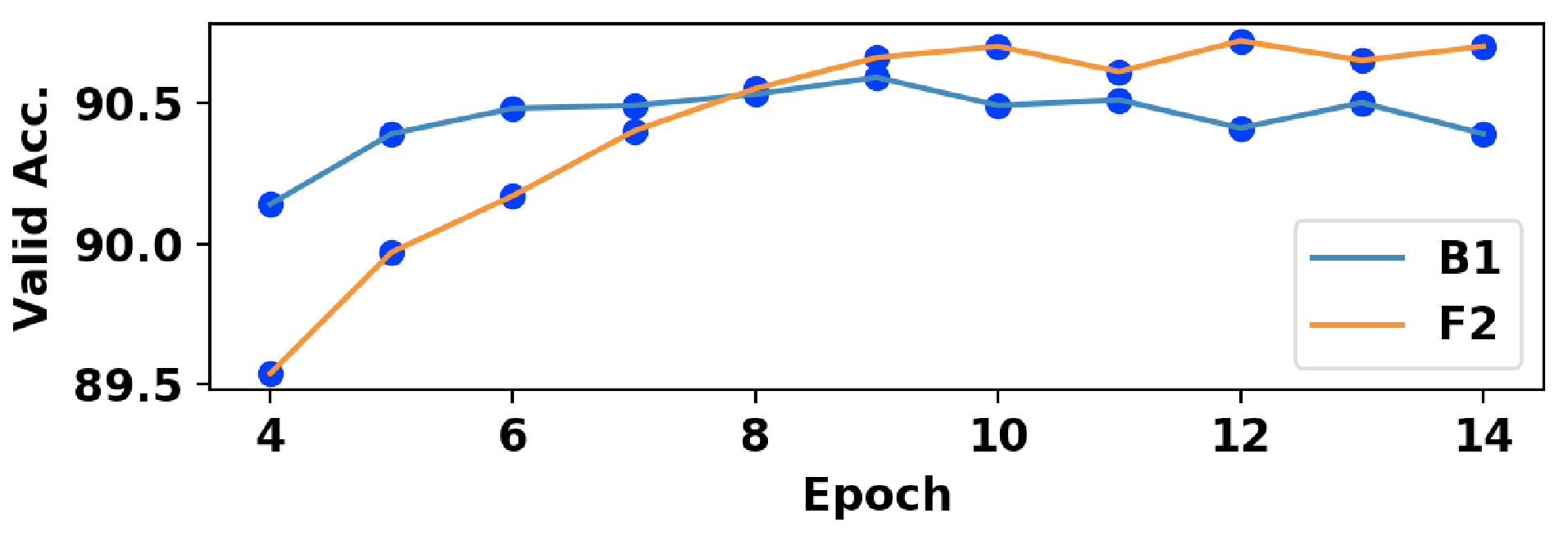}
 \caption{Word piece validation accuracy of ASR model in B1 and F2.}
  \label{fig:valid_acc}
\end{figure}

\textbf{Reducing overfitting} We further investigate the trend of word piece validation accuracy of the ASR model in \texttt{B1} and \texttt{F2}, as shown in Fig.~\ref{fig:valid_acc}. Compared to \texttt{B1}, \texttt{F2} learns more slowly and reaches a better local optimal. 
The learning objective of \texttt{F2} consists of both $L_{ASR}$ and accent-related regularizers (see Section \ref{sec:approach}). This observation corroborates the effectiveness of the regularization in our approach on reducing the risk of overfitting. It is worth noting that such benefit from the accent-related regularization in fine-tuning is not observed in supervised setting (see Table \ref{table:supervised}). 
The sufficient training data with high-quality transcriptions in supervised setting empowers the learning capability of the ASR model. As a result, the additional regularization might even weaken such learning capability. 

\section{Conclusion}
\label{sec:conclusion}
In this paper, we proposed \texttt{AIPNet}, a GAN-based pre-training network, for learning accent-invariant representations, aiming to build a unified speech recognition system that generalizes well across accents. As transcriptions are not needed in pre-training, \texttt{AIPNet} provides the flexibility of making use of many possible accent resources for which transcriptions are unavailable. Experiments have shown promising results on $9$ English accents compared to the baselines, especially in the case when transcriptions are not available in all accents. Experimental results have demonstrated the effectiveness of \texttt{AIPNet} on learning accent-invariance.



\bibliographystyle{IEEEbib}
\bibliography{refs}

\end{document}